\pdfoutput=1
\documentclass[letterpaper]{article} 
\usepackage[submission]{aaai24}  
\usepackage{times}  
\usepackage{helvet}  
\usepackage{courier}  
\usepackage[hyphens]{url}  
\usepackage{graphicx} 
\def\UrlFont{\rm}  
\usepackage{natbib}  
\usepackage{caption} 
\usepackage{newfloat}
\usepackage{listings}
\usepackage{times}

\usepackage{soul}
\usepackage{url}
\usepackage[utf8]{inputenc}
\usepackage{amsmath}
\usepackage{booktabs}
\usepackage{refcount}
\usepackage{amsfonts}
\usepackage{caption}
\usepackage{color, colortbl}
\usepackage{breqn}
\usepackage{bm}
\usepackage{enumerate}
\usepackage{subfigure}
\usepackage{subfloat}
\usepackage{multirow}
\usepackage{threeparttable}
\usepackage[table,xcdraw]{xcolor}
\usepackage{algorithm}
\usepackage{algorithmic}
\usepackage{tabularx}

\newtheorem{problem}{Problem}

\newtheorem{pro-stat}{Problem Definition}
\def\T{{\scriptscriptstyle\mathsf{T}}}

\newcommand{\optgnn}{\textsc{OPT-GNN}}

\newcommand{\hide}[1]{}

\DeclareCaptionStyle{ruled}{labelfont=normalfont,labelsep=colon,strut=off} 
\floatstyle{ruled}
\newfloat{listing}{tb}{lst}{}
\floatname{listing}{Listing}
\title{Optimal Propagation for Graph Neural Networks}
\author{}
\affiliations{}

\begin{document}
\setlength{\abovedisplayskip}{2.pt}
\setlength{\belowdisplayskip}{2.pt}
\setlength{\abovedisplayshortskip}{2.pt}
\setlength{\belowdisplayshortskip}{2.pt}

\maketitle

\begin{abstract}
Graph Neural Networks (GNNs) have achieved tremendous success in a variety of real-world applications by relying on the fixed graph data as input. However, the initial input graph might not be optimal in terms of specific downstream tasks, because of information scarcity, noise, adversarial attacks, or discrepancies between the distribution in graph topology, features, and groundtruth labels. In this paper, we propose a bi-level optimization approach for learning the optimal graph structure via directly learning the Personalized PageRank propagation matrix as well as the downstream semi-supervised node classification simultaneously. We also explore a low-rank approximation model for further reducing the time complexity. Empirical evaluations show the superior efficacy and robustness of the proposed model over all baseline methods. 
\end{abstract}

\section{Introduction}
Graph Neural Networks (GNNs) are powerful graph representation learning tools which aggregate information of neighborhood nodes with the given graph structural information and features. They have been applied to a wide range of real-world applications and have shown superiority over conventional methods. Representative applications include
recommender systems \cite{ying2018graph}, social networks \cite{li2016quint}, bioinformatics \cite{shang2019gamenet}, and many more. 

Most of the existing GNNs could be regarded as instantiations of message-passing GNNs, e.g., \cite{kipf2016semi,gilmer2017neural,velivckovic2017graph,wu2019simplifying, fu2021sdg} whose key idea is to pass and aggregate information of nodes/edges to neighboring nodes through graph structure in each layer of a multi-layered neural model. Among others, two recent advancements have exhibited significant performance improvement in representative graph learning tasks, namely: (\textit{M1}) predict-then-propagate, e.g., \cite{gasteiger2018predict} and propagate-then-predict methods, e.g., \cite{klicpera2018predict,chen2020scalable}, and (\textit{M2}) the structural learning-based methods, e.g., \cite{franceschi2019learning,jin2020graph,xu2022graph}. The key idea of \textit{M1} is to propagate the predictions/features via a pre-computed propagation matrix, and the key idea of \textit{M2} is to learn an updated graph adjacency matrix jointly with the downstream tasks. However, for \textit{M1}, calculating or estimating the accurate propagation matrix itself is costly in terms of computational complexity, since the normalized version of adjacency matrix should be calculated in each propagation. For \textit{M2}, although an updated adjacency matrix is learned, the adjacency matrix might not be as effective as the propagation matrix in certain graph learning tasks as suggested by methods in \textit{M1} \cite{gasteiger2018predict}. Message passing algorithm such as GCN is non-trivial to be expanded to a deeper architecture, due to oversmoothing and efficiency issues. Besides, more sophisticated methods using adjacency matrices such as GAT \cite{velivckovic2017graph} has been proved to be much slower than then propagation matrix-based method \cite{gasteiger2018predict}.

Generally speaking, the existing GNN-based models for various downstream graph learning tasks bear three limitations. First, most existing GNN models aggregate information to nodes based on a fixed given graph structure. However, there exists three types of correlated distributions in graph data, including the node/edge feature distribution, the topology (structure) distribution, and the label distribution. The divergence between the topology distribution and the label distribution are also studied in recent works as graph homophily and heterophily problems, which points out that GNNs' homophily assumption may fail to generalize them to graphs with
heterophily \cite{zhu2020beyond}. In the perspective of a given downstream task, the graph topology\footnote{
Features might also be not optimal, but we will mainly focus on the graph structure in this work.} might not be optimal for existing GNN models to adopt. The suboptimal topology might also because of other factors such as noise, data scarcity, etc. Second, for the recent predict-then-propagate or propagate-then-predict strategies, the pre-computed propagation matrix might not be optimal towards specific downstream tasks because of non-optimal adjacency matrix, from which the propagation matrix is calculated. The separation of propagation matrix computation and downstream task learning is another reason. Third, the existing GNN models are often vulnerable to adversarial attacks, such as node/edge deletion. In this paper, we try to answer the following research question: {\em how to develop a method that directly learns an optimal propagation matrix (instead of the adjacency matrix) towards different downstream tasks, and meanwhile improve the robustness when facing attacks?}

To tackle the aforementioned limitations, we propose \optgnn, 
a bi-level programming method. The lower level is aimed at learning an optimal graph propagation matrix by preserving the similarities/dissimilarities of nodes with identical/different labels. The upper level is the loss for the downstream task with the help of the learned Personalized PageRank (PPR) propagation matrix. The learning process of the PPR propagation matrix implicitly tailors the graph topology towards the downstream task, and enables better robustness against attacks. Furthermore, the direct learning of PPR propagation matrix provides flexibility and efficiency compared with the recent structure learning-based GNN models, whose key is to learn an updated adjacency matrix. 
To sum up, the main contributions of this paper are:
\begin{itemize}
\item \textbf{Formulation.} We introduce two novel optimization formulations based on PPR propagation matrix via bi-level programming to simultaneously learn the propagation matrix of the graph and optimize downstream tasks, such as node classification. 
\item \textbf{Algorithms.} We develop algorithms for solving the optimization problem of learning the propagation matrix, and further reduce the time complexity  by low-rank approximation.
\item \textbf{Evaluation.} We conduct extensive experiments on public node classification datasets to show the effectiveness. Results show that the proposed method surpasses state-of-the-art baselines in terms of accuracy and robustness. 
\end{itemize}

\section{Preliminaries and Problem Definition}\label{sec:problem}
In this section, we first unify the notations used throughout the paper. Then we start with the preliminaries of the Personalized PageRank (PPR) with its recent extensions in neural models and the recent structure learning-based methods, followed by the formal problem definition. 

\noindent \textbf{Notations. }
We use bold upper-case letters for matrices (e.g., $\mathbf{Q}$), bold lowercase letters for vectors (e.g., $\mathbf{p}$), lowercase letters (e.g., $\alpha$) for scalars, and the superscript $^{T}$ for matrix and vector transpose (e.g., $\mathbf{X}^{\top}$ is the transpose of  $\mathbf{X}$).
We use $\mathbf{Q}(a_i,a_j)$ to denote the entry at $(a_i, a_j)$ of matrix $\mathbf{Q}$, and $\mathbf{p}[a_i]$ to denote the $a_i$-th entry of the vector $\mathbf{p}$.

We represent a graph $G = (\mathcal{V}, \mathcal{E})$  with node set $\mathcal{V}$ and edge set $\mathcal{E}$. $n$ and $m$ denote the numbers of nodes and edges. Each node $v$ is associated with a feature vector of dimension $d$, so the feature matrix $\mathbf{X} \in  \mathbb{R}^{n \times d}$. $\mathbf{A}$ is an $n \times n$ adjacency matrix. $\mathbf{Y}=\{y_1, y_2,...,y_n\}$ is the groundtruth label distribution. $\tilde{\mathbf{A}}=\mathbf{D}^{-\frac{1}{2}} \mathbf{AD}^{-\frac{1}{2}}$ is the symmetrically normalized adjacency matrix, where $\mathbf{D}$ is the diagonal degree matrix of $\mathbf{A}$. $\mathcal{V}^{(y)}$ represents the node set with labels. Nodes with the label $k$ are in the set $\mathcal{C}_k$, nodes with labels different from $k$ are in the set $\mathcal{V}^{(y)}\backslash \mathcal{C}_k$. Among nodes in each mini-batch, those belong to the set $\mathcal{C}_k$ are denoted as ${a}_{j}$, and those in $\mathcal{V}^{(y)}\backslash \mathcal{C}_k$ are labeled as ${b}_{j}$. Numbers of $a_{j}$ and $b_{j}$ are $|\mathcal{P}|$ and $|\mathcal{N}|$ respectively. The final prediction result is defined as $\mathbf{Z} \in  \mathbb{R}^{n \times K}$. $L$ is the layer of the GNN model. $K$ is the number of classes.

\subsection{Preliminaries}
\noindent \textbf{Personalized Propagation of Neural Predictions (PPNP).}
PPNP and its approximate version APPNP with power iteration belong to predict-then-propagate methods. Procedures for these methods are: $(1)$ predicting over the original node features $\mathbf{H}^{(0)}=f_{\theta}(\mathbf{X})$, where $f_{\theta}\left(\cdot\right)$ is a neural network to generate predictions with parameter set $\theta$; (2) propagating  the  prediction results with the idea of Personalized PageRank in each layer $\mathbf{H}^{(l+1)}=(1-\alpha)\tilde{\mathbf{A}}\mathbf{H}^{(l)}+\alpha\mathbf{H}^{(0)}$, where $0<\alpha <1$; $(3)$ If the number of stacked layers is a finite number $L$, the model is named as APPNP. After the $k$-th propagation step, the resulting predictions are $\mathbf{H}^{(l)}=\left((1-\alpha)^{l} \tilde{\mathbf{A}}^{l}+\alpha \sum_{i=0}^{l-1}(1-\alpha)^{i} \tilde{\mathbf{A}}\right) \mathbf{H}^{(0)}$, which aims to ensure high efficiency with a trade-off between power iteration number and accuracy;
$(4)$ If the number of stacked layers is infinite, which is to take the limit $L \rightarrow \infty$, the left term tends to 0 and the right term becomes a geometric series. Also, since $\tilde{\mathbf{A}}$ is symmetrically normalized, det($\tilde{\mathbf{A}}$) $\leq$ 1. Finally, we have $\mathbf{H}^{(\infty)}=\alpha\left(\mathbf{I}_{n}-(1-\alpha) \tilde{\mathbf{A}}\right)^{-1} \mathbf{H}^{(0)}$ which is equivalent to the closed-form solution of PPR and named as PPNP. 
Thus, we obtain the final predictions of downstream tasks as  $\textrm{softmax}(\alpha(\mathbf{I}-(1-\alpha)\tilde{\mathbf{A}})^{-1}\mathbf{H}^{(0)})$, and it is named as PPNP:
\begin{equation}
    \mathbf{Z}_{\mathrm{PPNP}}=\operatorname{softmax}\left(\alpha
     \mathbf{Q} f_{\theta}(\mathbf{X})\right)
\end{equation}
in which we define $\mathbf{Q}=(\mathbf{I}-(1-\alpha)\tilde{\mathbf{A}})^{-1}$ as the PPR propagation matrix. It measures the proximity of any two nodes in the graph. For example, $\mathbf{Q}(a_i,a_j)$ and $\mathbf{Q}(a_i,b_j)$ reflect how much proximity between nodes $a_i$, $a_j$ and $a_i$, $b_j$ respectively. In node classification problems, it is often expected that nodes with the same label are more proximate than those with different labels, so $\mathbf{Q}(a_i,a_j)$ is larger than $\mathbf{Q}(a_i,b_j)$. 

In this work, we will study how to obtain a {\em learned} PPR propagation matrix $\mathbf{Q}_{s}$ to substitute $\mathbf{Q}$ to make the graph topology, the feature distribution and the label distribution more aligned with each other to achieve a better performance on node classification and robustness performance.

\noindent \textbf{Structure Learning-based Methods.} Most of existing structure learning-based methods \cite{franceschi2019learning,jin2020graph,xu2022graph} try to learn an optimal {\em adjacency matrix} by adding or subtracting edges. Instead, we propose to learn an optimal {\em propagation matrix} with two advantages, including $(1)$ efficiency: in terms of propagation matrix-based methods, we do not need to calculate $\mathbf{Q}_{s}$ from the adjacency matrix $\mathbf{A}_{s}$ which contains multiple propagation steps, and then conducting backpropagation in a reverse step; and $(2)$ flexibility: 
the adjacency matrix must be kept sparse to avoid costly matrix multiplication, while our strategy (directly learning $\mathbf{Q}_s$) is more flexible than the method of first learning $\mathbf{A}_s$ and then computing $\mathbf{Q}_s$. The task setup is to directly learn an optimal PPR propagation matrix $\mathbf{Q_s}$ to substitute $\mathbf{Q}$ and realize higher downstream classification accuracy on the upper level. 

\subsection{Problem Definition}

The problem that this paper studies is formally defined as follows:

\begin{problem} {\textbf{Optimal Propagation Matrix Learning}}

\textbf{Given:} The input graph $G=(\mathcal{V}, \mathcal{E})$, the initial node feature matrix $\mathbf{X}$, the groundtruth labels $\mathbf{Y}$ in the training set of a given downstream task;

\textbf{Output:} The optimal propagation matrix $\mathbf{Q}_s$ for learning the downstream task, and the predictions on testing set of the given task. 


\end{problem}

\section{Proposed Model}\label{sec:method}
In this section, we first introduce our PPR propagation matrix learning-based bi-level programming model. Then, we propose an approximate algorithm based on rank-one perturbation assumption to reduce the time complexity. After that, we present the computational complexity analysis. 

\subsection{The \optgnn\ Model}
\noindent \textbf{Bi-level Programming Framework.}
In our method, the upper-level aims to minimize the node classification loss based on the predicted labels via a GNN model, whose inputs include the feature matrix $\mathbf{X}$ of nodes, classification labels $\mathbf{Y}$, and the learned propagation matrix $\mathbf{Q}_s$ from the lower-level optimization formulation. In the lower-level optimization, the goal is to directly learn a propagation matrix $\mathbf{Q}_s$ which encodes feature and class information for the upper level. To be specific, the proposed bi-level programming formulation is:
\begin{multline}\label{eq:framework}
\begin{array}{l}
\min _{\theta} \mathcal{L}_{u}\left(\mathbf{Q}_{s}, \mathbf{X}, \mathbf{Y}, \theta\right) \\
\text { s.t., }\mathbf{Q}_{s} = \arg \min_{\mathbf{Q}_s} \mathcal{L}_{l}\left(\mathbf{Q}_{s}, \mathbf{Q}, \mathbf{X}, \mathbf{Y}\right)
\end{array}
\end{multline}

where $\mathcal{L}_u()$ and $\mathcal{L}_l()$ denote upper and lower optimization loss, respectively. $\mathbf{Q}$ is the propagation matrix of the initial adjacency matrix, $\mathbf{Q}_s$ is the learned propagation matrix. The lower level objective is to learn an optimal graph whose corresponding propagation matrix can be used for the upper level optimization. Compared to many methods that work on adjacency matrix update \cite{li2016quint,jin2020graph,xu2022graph} 
, we directly focus on optimizing the propagation matrix instead of learning its adjacency matrix first and calculating the corresponding propagation matrix. This brings two important advantages. First, it will reduce the time complexity. Second, since the two levels of optimization formulation do not share parameters, we can solve Eq. \eqref{eq:framework} in a separate way without the calculation of hypergradient as many existing bi-level programming methods. This will simplify the gradient computation as well as avoiding hypergradient approximation errors.

Using the PPR propagation matrix, and multi-class classification as an instantiation, the upper level optimization is:
\begin{equation} \label{eq:upper_loss}
    \min_{\theta} \mathcal{L}_c(\operatorname{softmax}(\alpha \mathbf{Q}_s f_{\theta}(\mathbf{X}))
\end{equation}
where $\mathcal{L}_c()$ is a regularized cross-entropy loss, and $f_{\theta}()$ is a neural function with learnable parameter $\theta$, which takes the node feature $\mathbf{X}$ as input. The lower level optimization is:
\begin{dmath*}\nonumber
    \min_{\mathbf{Q}_s} ||\mathbf{Q}_s - \mathbf{Q}||^2_F +  tr(\mathbf{X}^\T\mathbf{Q}_s\mathbf{X}), ~~\\
    s.t. \mathbf{Q}_s(a_i, a_j) > \mathbf{Q}_s(a_i, b_j), \forall a_i, a_j\in \mathcal{C}_k, b_j \in \mathcal{V}^{(y)}\backslash \mathcal{C}_k.
\end{dmath*}

The first term is the norm of the difference between the learned $\mathbf{Q}_s$ and $\mathbf{Q}$. The second term  $tr(\mathbf{X}^\T\mathbf{Q}_s\mathbf{X})$ controls the smoothness of features in the learned graph propagation matrix, aiming to reduce the divergence between the learned propagation matrix with the node features. The motivation of the constraint is to eliminate the discrepancy of the distribution of the graph topology and labels. Specifically, it tries to increase the entry of propagation matrix if two nodes are from the same class, and otherwise reducing it. The weighting parameters are omitted here for clarity. 

However, in the above equation, the constraint of entries in the propagation matrix is too hard, so we change it to a penalization term to involve some relaxation. After modification, the soft version of lower level optimization is:
\begin{multline}\label{eq:lower_opt1}
\min _{\mathbf{Q}_{s}} \left\|\mathbf{Q}_{s}-\mathbf{Q}\right\|_{F}^{2}+\epsilon \operatorname{tr}\left(\mathbf{X}^{\top} \mathbf{Q}_{s} \mathbf{X}\right)+\\
\sum_{a_{i}, a_{j} \in \mathcal{C}_{k} \atop b_{j} \in \mathcal{V}^{(y)} \backslash \mathcal{C}_{k}}
 c g\left(\mathbf{Q}_{s}\left(a_{i}, b_{j}\right)-\mathbf{Q}_{s}\left(a_{i}, a_{j}\right)\right)
\end{multline}
where $\epsilon$ and $c$ are trade-off parameters that balance among terms. We use the variant of the sigmoid function as $g(\cdot)$:
$$
g(x)=\frac{1}{1+\exp (-x / b)}.
$$
with $b$ as a parameter to map the difference of the graph topology and the label distribution between 0 and 1.
When $\mathbf{Q}_{s}\left(a_{i}, b_{j}\right)<\mathbf{Q}_{s}\left(a_{i}, a_{j}\right)$, which is what we expect, we let $g(\cdot)=0$. We refer to our model with lower-level optimization of Eq. \eqref{eq:lower_opt1} as the \optgnn\ model. 

\subsection{Optimization Solution}
The lower level optimization function is non-convex, so we use gradient descent-based methods to calculate the derivative of the lower level loss function and update along the direction of the fastest gradient descent. Here, we show steps to obtain the gradient of Eq. (\ref{eq:lower_opt1}). $\mathbf{Q}_{s}\left(a_{i}, b_{j}\right)$ can be written as:
$$\mathbf{Q}_{s}\left(a_{i}, b_{j}\right)=\mathbf{e}_{{a_j}}^{\top} \mathbf{Q}_{\mathrm{s}} \mathbf{e}_{b_j}$$
where $\mathbf{e}_{a_{i}}$ and $\mathbf{e}_{b_{j}}$ are n-dimensional vectors with 1 for the $a_j$-th and $b_j$-th entries, and 0 for other entries. Then, we calculate the derivative of $\mathbf{Q}_{s}\left(a_{i}, b_{j}\right)$ with respect to $\mathbf{Q}_{s}$, 
$$
\frac{\partial \mathbf{Q}_{s}\left(a_{i}, b_{j}\right)}{\partial \mathbf{Q}_{s}}=\frac{\partial \mathbf{e}_{a_{i}}^{\top} \mathbf{Q}_{\mathbf{s}} \mathbf{e}_{b_{j}}}{\partial \mathbf{Q}_{s}}=\mathbf{e}_{a_{i}} \mathbf{e}_{b_{j}}^{\top},
$$
Therefore, the derivative for Eq. \eqref{eq:lower_opt1} is:
\begin{multline}\label{eq:lower_de}
\frac{\partial \mathcal{L}\left(\mathbf{Q}_{s}\right)}{\partial \mathbf{Q}_{s}}
=2\left(\mathbf{Q}_{s}-\mathbf{Q}\right)+\epsilon \mathbf{X X}^{\top}+\\
\sum_{a_{i}, a_{j} \in \mathcal{C}_{k} \atop b_{j} \in \mathcal{V}^{(y)} \backslash \mathcal{C}_{k}} c \frac{\partial g\left(d_{y x}\right)}{\partial d_{y x}} \mathbf{e}_{a_{i}}\left(\mathbf{e}_{b_{j}}-\mathbf{e}_{a_{j}}\right)^{\top}
\end{multline} 
where $d_{y x}=\mathbf{Q}_{s}(a_{i}, b_{j})-\mathbf{Q}_{s}(a_{i}, a_{j})$ and $\frac{\partial g(d_{y x})}{\partial d_{y x}}=\frac{1}{b} g(d_{y x})(1-g(d_{y x}))$.
The corresponding algorithm is shown in Algorithm \ref{alg:algorithm1}. 

With the learned $\mathbf{Q_s}$, we feed it into a two-layer GNN to train the upper level optimization model and solve the semi-supervised node classification problem as shown in Eq. \eqref{eq:upper_loss}. The loss function $\mathcal{L}_{c}$ is a regularized cross-entropy loss: 
$$\mathcal{L}_{c}=-\sum_{i}^{n} \sum_{j}^{K} \mathbf{Y}(i, j) \ln \mathbf{Z}(i, j)+\lambda\left\|\mathbf{\Theta}_{1}\right\|^{2}
$$
where $\mathbf{\Theta}_{1}$ denotes parameters for the upper level, $\mathbf{Z}$ is the prediction result, and $\lambda$ is the regularization coefficient.

\begin{algorithm}[!htb]
\caption{The lower-level of \optgnn} 
\label{alg:algorithm1}
\begin{algorithmic}[1]
\REQUIRE The initial PPR propagation matrix $\mathbf{Q}$;\\
Nodes $a_i$ with label $k$;\\
Nodes $a_j$ from the set $\mathcal{C}_k$ and $b_j$ from the set $\mathcal{V}^{(y)} \backslash\mathcal{C}_k $;\\
Parameters  $\epsilon$, $c$, the step size $\eta$ and the maximum iteration number $T$.

\ENSURE The learned PPR propagation matrix $\mathbf{Q}_s$. 
\STATE Initialize $\mathbf{Q}_{s}=\mathbf{Q}$; 
\WHILE{not converged or iteration $\le$ $T$} 
\FOR{each node $a_{j}$ in the class $k$}
\FOR{each node $b_{j}$ not in the class $k$}
\STATE Compute $\frac{\partial g\left(d_{y x}\right)}{\partial d_{y x}} \mathbf{e}_{a_{i}}\left(\mathbf{e}_{b_{j}}-\mathbf{e}_{a_{j}}\right)^{\T}$;
\ENDFOR
\ENDFOR
\STATE Compute the derivative $\frac{\partial \mathcal{L}\left(\mathbf{Q}_{s}\right)}{\partial \mathbf{Q}_{s}}$ by Eq.~(\ref{eq:lower_de});
\STATE Update $\mathbf{Q}_{s}=\mathbf{Q}_{s}-\eta\cdot \frac{\partial \mathcal{L}\left(\mathbf{Q}_{s}\right)}{\partial \mathbf{Q}_{s}}$;
\ENDWHILE
\RETURN The learned optimized PPR propagation matrix;
\end{algorithmic}
\end{algorithm}

\subsection{Approximate Optimal Propagation Matrix Learning (AOPT-GNN)}
\label{sec:AOPT-GNN}
Here, we present AOPT-GNN to reduce the time complexity. The key idea is that the lower level optimization problem of learning an optimal propagation matrix can be further approximated via a low-rank form (e.g., rank-one).

As suggested by \cite{li2016quint}, the approximate optimal propagation matrix is a low-rank perturbation of the original PPR propagation matrix. We assume that $\mathbf{Q}_{s}$ is the rank-one perturbation of $\mathbf{Q}$ as $\mathbf{Q_s}= \mathbf{Q}+ \mathbf{p q}^{\top}$.

After adding weighting parameters, the lower level loss can be written as:
\begin{multline}\label{eq:lower_opt_fast}
\min _{\mathbf{p}, \mathbf{q}} \left\|\mathbf{p q}^{\top}\right\|_{F}^{2}+\beta\left(\|\mathbf{p}\|^{2}+\|\mathbf{q}\|^{2}\right)+\gamma \operatorname{tr}\left(\mathbf{X}^{\top} \mathbf{p q}^{\top} \mathbf{X}\right)+\\
\sum_{a_{i}, a_{j} \in \mathcal{C}_{k} \atop b_{j} \in \mathcal{V}^{(y)} \backslash \mathcal{C}_{k}} c g\left(\mathbf{Q}_{s}\left(a_{i}, b_{j}\right)-\mathbf{Q}_{s}\left(a_{i}, a_{j}\right)\right)
\end{multline}
We first keep the vector $\mathbf{q}$ as a constant vector. The derivative of Eq. (\ref{eq:lower_opt_fast}) with respect to the vector $\mathbf{p}$ is:
\begin{multline} \label{eq:lower_opt_fast_diff}
\frac{\partial \mathcal{L}(\mathbf{p})}{\partial \mathbf{p}}=2 \mathbf{p q}^{\top} \mathbf{q}+2 \beta \mathbf{p}+\gamma \mathbf{X X}^{\top} \mathbf{q}+\\ \sum_{a_{i}, a_{j} \in \mathcal{C}_{k} \atop b_{j} \in \mathcal{V}^{(y)} \backslash \mathcal{C}_{k}} c \frac{\partial g\left(d_{y x}\right)}{\partial d_{y x}}  \mathbf{q}[{a}_{i}]\left(\mathbf{e}_{b_j}-\mathbf{e}_{a_j}\right)
\end{multline}
where $\mathbf{q}[a_{i}]$ is the $a_i$-th entry of vector $\mathbf{q}$. The derivative of Eq.~(\ref{eq:lower_opt_fast}) with respect to the vector $\mathbf{q}$ can be computed in a similar way with vectors $\mathbf{p}$ and $\mathbf{q}$ interchanged. The lower level algorithm with low-rank approximation is shown in Algorithm \ref{alg:algorithm2}. The upper level loss is the same as \optgnn.

\begin{algorithm}[!htb]
\caption{The lower-level of A\optgnn} 
\label{alg:algorithm2}
\begin{algorithmic}[1]
\REQUIRE The initial PPR propagation matrix $\mathbf{Q}$;\\
nodes $a_i$ with label $k$;\\
nodes $a_j$ from the set $\mathcal{C}_k$ and $b_j$ from the set $\mathcal{V}^{(y)} \backslash\mathcal{C}_k$;\\
parameters  $\beta,\gamma, c$, the step size $\eta$ and the maximum iteration number $T$.\\
\ENSURE The rank-one perturbation vectors $\mathbf{p}$ and $\mathbf{q}$.
\STATE Initialize $\mathbf{p}$ and $\mathbf{q}$; 
\WHILE{not converged or iteration $\le$ $T$}
\FOR{each node $a_{j}$ in the class $k$}
\FOR{each node $b_{j}$ not in the class $k$}
\STATE Compute $c \frac{\partial g\left(d_{y x}\right)}{\partial d_{y x}} \mathbf{q}{[a_{i]}}\left(\mathbf{e}_{b_{j}}-\mathbf{e}_{a_{j}}\right)$;
\STATE Compute $c \frac{\partial g\left(d_{y x}\right)}{\partial d_{y x}} \mathbf{p}{[a_{i}]}\left(\mathbf{e}_{b_{j}}-\mathbf{e}_{a_{j}}\right)$;
\ENDFOR
\ENDFOR
\STATE Compute the derivative $\frac{\partial \mathcal{L}(\mathbf{p})}{\partial \mathbf{p}}$ by Eq.~ (\ref{eq:lower_opt_fast_diff});
\STATE Compute the derivative $\frac{\partial \mathcal{L}(\mathbf{q})}{\partial \mathbf{q}}$;
\STATE Update $\mathbf{p}=\mathbf{p}-\eta\cdot \frac{\partial \mathcal{L}\left(\mathbf{p}\right)}{\partial \mathbf{p}}$;
\STATE Update $\mathbf{q}=\mathbf{q}-\eta\cdot \frac{\partial \mathcal{L}\left(\mathbf{q}\right)}{\partial \mathbf{q}}$;
\ENDWHILE
\RETURN The learned vectors $\mathbf{p}$ and $\mathbf{q}$.
\end{algorithmic}
\end{algorithm}

 
\subsection{Complexity Analysis}
\label{subsec:Complexity_Analysis}
The purpose of our approximate \optgnn\ is to reduce the time complexity. Here, we give some theoretical analysis about the time complexity of our base algorithm \optgnn\ and approximate algorithm A\optgnn. For lower level training, the process from line 3 to 5 of Algorithm~\ref{alg:algorithm1} repeats $ |\mathcal{P}||\mathcal{N}|$ 
times. The time complexity of line 5 is $O\left(n^2\right)$, and that of the whole algorithm is 
$O\left(T_{1} |\mathcal{P}| |\mathcal{N}|n^2\right)$, where $T_1$ is the number of iterations to convergence at the lower level, and $n$ is the node number of the graph. In AOPT-GNN, the inner time complexity of the algorithm is simplified to $O\left(n\right)$, and the whole time complexity of the lower level is reduced to $O\left(T_1 |\mathcal{P}| |\mathcal{N}|n\right)$, which realizes a linear time complexity in the updating process. Since $\mathbf{X} \in \mathcal{R}^{n \times d}$, and $d$ is the dimension of the feature vector, the upper level time complexity of \optgnn\ and A\optgnn\ is $O\left(T_{2} L n d^2\right)$, where $L$ is the layer of the GNN, and $T_{2}$ is the number of iterations to convergence for training the GNN.
It is worth mentioning that the original propagation matrix $\mathbf{Q}$ is only calculated once before the training process, which is not involved in the training process. Also, the computation is trivial compared to the training time. Therefore, it is not discussed in the complexity analysis.


\section{Experiments}\label{sec:experiments}
In this section, we evaluate the effectiveness, robustness, and the efficiency of the proposed model on the semi-supervised node classification task. 

\subsection{Experiment Setup}
\label{subsec:experiment_setup}
\noindent \textbf{Datasets.} We evaluate performances of \optgnn\ and A\optgnn\ on datasets Cora \cite{mccallum2000automating}, Cora-ML \cite{bojchevski2017deep}, Citeseer \cite{sen2008collective}, Polblogs \cite{adamic2005political} and Pubmed \cite{namata2012query}. Cora, Cora-ML and Citeseer are citation graphs with nodes representing papers and edges representing citations between them. Polblogs is a social network dataset, where blogs are nodes and their cross-references are edges. It is commonly used in adversarial settings. Since Polblogs do not have node features, an $n \times n$ identity matrix is used as the feature matrix. Statistics of the datasets used in effectiveness experiments are summarized in Table \ref{tab:Dataset_statistics}. The datasets are split into visible and invisible sets. In the visible set, $20$ nodes per class are used for training each epoch, and $500$ nodes are used for validation. Invisible set is used as the test set, which are $1,000$ nodes. 

\begin{table}[h]
\centering
{\fontsize{9.5pt}{10pt} \selectfont
\caption{Datasets statistics and segmentation.}
\label{tab:Dataset_statistics}
\setlength{\tabcolsep}{3pt}{
\begin{tabular}{@{}cccccccc@{}}
\toprule
Dataset  & Nodes & Edges  & Features & Class & Train &Val & Test \\ \midrule
Cora & 2,708  & 5,429   & 1,433     & 7       & 140 & 500 &1,000   \\
Cora-ML & 2,810  & 7,981   & 2,879     & 7       & 140 & 500 &1,000   \\
Citeseer & 3,327  & 4,372   & 3,703     & 6       & 120 & 500 &1000     \\
Polblogs & 1,222 & 16,714 & N/A      & 2       &   60  & 300 &500        \\
Pubmed & 19,717 & 44,338 &  500  &  3    &  60   & 500 &1,000       \\
 \bottomrule
\end{tabular}}}
\end{table}

\noindent \textbf{Implementation Details.}
Experiments are repeated with five different data segmentation and each with five different random seeds, unless otherwise specified In robustness experiments, we adopt the data segmentation the same as \cite{jin2020graph} with $10\%$ nodes used for training and $80\%$ nodes for testing. 
AOPT-GNN requires the initialization of vectors $\mathbf{p}$ and $\mathbf{q}$. We let $\mathbf{p}[b_j] = \mathbf{Q}\left(a_i,b_j\right)$, $\mathbf{p}[a_j] = - \mathbf{Q}\left(a_i,a_j\right)$, $\mathbf{p}[a_i] = \mathbf{Q}\left(a_i,a_i\right)$, and 0 for other entries. $\mathbf{q}$ is an {\em n}-dimensional zero vector. This setting ensures that $\mathbf{Q}_s = \mathbf{Q}$ at the initialization stage in Algorithm~\ref{alg:algorithm2}. The computational resources are NVIDIA Tesla K80 GPU and NVIDIA 3090Ti GPU.

\noindent \textbf{Baseline Comparison.} \label{subsec:effectiveness}
We compare our methods with the following baseline models, which are state-of-the-art methods or representatives in (1) pure GNNs: GCN \cite{kipf2016semi}, SGC \cite{wu2019simplifying}, PPNP \cite{klicpera2018predict} and GAT \cite{velivckovic2017graph}; (2) heuristics-based modification methods: Jaccard \cite{wu2019adversarial}, SVD \cite{entezari2020all}; (3) learning-based methods: LDS \cite{franceschi2019learning}, Pro-GNN \cite{jin2020graph}, GEN \cite{wang2021graph}, PTDNet \cite{luo2021learning} and GASOLINE \cite{xu2022graph} (Due to different data segmentation, please refer to Table \ref{tab:attack} and the supplementary material for the comparison under no attack). 
\begin{table}[h]
\centering
{\fontsize{9pt}{11pt} \selectfont
\begin{threeparttable}
\setlength{\tabcolsep}{1pt} 
\caption{Mean accuracy comparison among baselines.}\label{tab:accuracy}
\begin{tabularx}{8.3cm}{@{}cccccc@{}}
\toprule
Dataset & Cora & Cora-ML & Citeseer & Polblogs & Pubmed\\ 
\midrule
GCN & 81.7$\pm$0.8 & 81.6$\pm$0.7 & 70.6$\pm$2.6 & 92.4$\pm$0.6 &78.6$\pm$0.5\\
GAT & 82.3$\pm$1.0 & 83.2$\pm$0.1 & 71.9$\pm$0.7 & \underline{94.7$\pm$0.3 } &77.4$\pm$0.4\\
SGC & 80.9$\pm$0.4 & 81.0$\pm$0.1 & 71.9$\pm$0.1 &92.8$\pm$1.1 &77.1$\pm$2.6\\
PPNP & 83.3$\pm$1.0 &  \underline{84.4$\pm$0.8}& \underline{75.0$\pm$0.7} &93.8$\pm$0.8  &79.7$\pm$0.4 \\ 
\midrule
Jaccard & 81.6$\pm$0.7 & 81.0$\pm$0.7 &71.3$\pm$2.2 & N/A\tnote{1}& -\tnote{2}\\ 
SVD & 81.6$\pm$0.7 &80.6$\pm$1.1 &61.4$\pm$2.3 & 89.8$\pm$1.8&-\tnote{2}\\ 
\midrule
LDS & 82.5$\pm$0.9 & 83.1$\pm$1.7&72.7$\pm$0.7&93.8$\pm$0.8  &78.2$\pm$1.8\\ 
Pro-GNN & 80.9$\pm$0.9 &83.0$\pm$0.2&73.3$\pm$ 0.7&N/A\tnote{1}  &78.0$\pm$0.8\\
GEN & 83.6$\pm$0.4 &-&73.8$\pm$ 0.6&- &\textbf{80.9$\pm$0.9}\\
PTDNet &\textbf{84.4$\pm$2.3}&-&73.7$\pm$ 3.1& -&\underline{79.8$\pm$2.4}\\
\midrule
\optgnn\  &  \underline{84.1$\pm$0.7}& \textbf{85.2$\pm$0.4} &\textbf{75.6$\pm$0.3 }& \textbf{94.8$\pm$0.5} & \textbf{81.1 $\pm$ 0.9} \\
A\optgnn\  & 84.0$\pm$0.7& \textbf{85.2$\pm$0.8} &\textbf{75.5$\pm$0.3 }& \textbf{94.8$\pm$0.4}  & \textbf{80.9$\pm$0.3}\\ 
 \bottomrule
\end{tabularx}
\begin{tablenotes} 
\footnotesize
\item[1] N/A: Jaccard and Pro-GNN cannot be directly used to datasets which do not have node features.
\item[2] Due to different data segmentation, please refer to supplementary material for the comparison under no attack.
\end{tablenotes}            
    \end{threeparttable} }
\end{table}

\noindent \textbf{Hyper-parameter Settings.} \label{sec:Hyper-parameter_Settings}
For each iteration, after randomly selecting a node $a_{i}$, we pick $50$ nodes $a_{j}$ and $50$ nodes $b_{j}$ from visible nodes for Cora, Cora-ML, Citeseer and Pubmed. For Polblogs, the number changes to $30$. The maximum iteration of the lower level is set to $200$ for Cora, Cora-ML, Citeseer and Pubmed, and $100$ for Polblogs. At the lower level, $\eta$ controls how much to update the propagation matrix, which is $0.01$ in all experiments. 
$b$ used in the function of the function $g(\cdot)$ is set to $0.01$.
As for the higher level, which is training of GNN, hidden units and graph neural network layers are $64$ and $2$, which are the same as \cite{klicpera2018predict}. $L_2$ regularization with $\lambda= 0.005$ is applied to the weights of the first layer, and dropout is set to $0.1$ for all experiments of our models.  
For baselines, learning rate is $0.01$ for GCN, GAT, Pro-GNN, and $0.1$ for SGC and PPNP. For LDS, the inner and outer learning rates are $0.02$ and $1$ \cite{franceschi2019learning}. Dropout rate is $0.6$ for GAT and $0.5$ for others. Hidden layers are all kept the same as their original papers. Layer number is set to $2$ in all models for fair comparison. For those methods based on PPR propagation matrix, $\alpha$ is $0.1$. For those without publicly available code, we report the experimental result from original papers with the same division of datasets.

\subsection{Effectiveness Results}
{\em A - Node Classification Performance.} The node classification performance of our model as well as baselines are shown in Table \ref{tab:accuracy} measured by the mean accuracy with standard deviation. To reduce the error, datasets are divided five times with different seeds. As we can see, both the proposed \optgnn\ and A\optgnn\ models outperform all different categories of baseline methods on Cora-ML, Citeseer, Polblogs and a compepitive result on Cora and Pubmed, showing the superiority without attack. The learned PPR propagation matrix indeed helps with the downstream node classification tasks.

\noindent {\em B - Robustness Performance.} The core purpose of this paper is to learn an optimal propagation matrix for graph, not to focus on resistance to attacks, while recent work demonstrates that adversarial attacks on graphs tend to connect nodes with distinct features \cite{wu2019adversarial}, which means to introduce more noises to the graph structure. Therefore, a method of optimizing the graph structure is theoretically able to correct the attacked structure to the optimum. We present the robustness performance of the proposed model in Table~\ref{tab:attack} 
in terms of mean accuracy and standard deviation of OPT-GNN and AOPT-GNN under \emph{metattack}. We report the best result of variants of all methods. Please refer to the supplementary material for the complete form. In the face of attack or interference to the graph, our model shows significant better robustness than PPNP \cite{gasteiger2018predict}, with a fixed PPR propagation matrix. Besides, although the main target of this paper is not to defense attacks, \optgnn can achieve a competitive or better performance compared to recent graph structure learning models Pro-GNN \cite{jin2020graph}, and GASOLINE \cite{xu2022graph} for attack defense.

\begin{table}[H]
\centering
{\fontsize{9pt}{11pt} \selectfont
\setlength{\tabcolsep}{1pt}
\caption{Performance (Accuracy$\pm$Std) under \emph{metattack}. }
\label{tab:attack}
\begin{tabularx}{8.3cm}{@{}cccccc@{}}
\toprule
Dataset & Ptb (\%) & PPNP& 
Pro-GNN &
GASOLINE &
(A)\optgnn \\ \midrule
\multirow{6}{*}{Cora} 
& 0 & 84.0$\pm$0.4 & 83.0$\pm$0.2&85.2$\pm$0.2&\textbf{85.8$\pm$0.7} \\
& 5  & 74.1$\pm$0.7 &82.3$\pm$0.4&77.4$\pm$0.5& \textbf{83.4$\pm$0.6}  \\
& 10 & 65.2$\pm$0.4 &79.0$\pm$0.6&70.8$\pm$0.5& \textbf{81.3$\pm$0.4}  \\
& 15 & 58.2$\pm$1.1 &76.4$\pm$1.3&67.1$\pm$0.8&  \textbf{78.7$\pm$1.7} \\
& 20 & 51.7$\pm$0.7 &\textbf{73.3$\pm$1.6}&62.5$\pm$0.5&72.5$\pm$2.5 \\
& 25 &  47.0$\pm$0.7 &\textbf{69.7$\pm$1.7}&57.3$\pm$0.6&64.5$\pm$1.5  \\\midrule
\multirow{6}{*}{Citeseer} 
& 0  & 71.8$\pm$0.4 &73.3$\pm$0.7&74.7$\pm$0.2&\textbf{75.5$\pm$0.7}  \\
& 5  & 67.6$\pm$0.9 &72.9$\pm$0.6&69.6$\pm$0.7&\textbf{74.6$\pm$0.6}   \\
& 10 &  61.8$\pm$0.8 &72.5$\pm$0.8&66.3$\pm$1.0&\textbf{73.9$\pm$0.5}  \\
& 15 & 54.1$\pm$1.2 &72.0$\pm$1.1&59.3$\pm$1.1& \textbf{74.1$\pm$0.6} \\
& 20 & 51.0$\pm$1.2 &70.0$\pm$2.3&56.5$\pm$0.9&  \textbf{73.8$\pm$0.6}\\
& 25 &49.4$\pm$2.2  &\textbf{69.0$\pm$2.8}&56.5$\pm$0.8&  65.6$\pm$1.0 \\
& 25 & 57.0$\pm$3.6 & 63.3$\pm$4.4& \textbf{89.9$\pm$0.5}& 67.1$\pm$2.9\\ \bottomrule
\end{tabularx}}
\end{table}

\subsection{Ablation Study}
\begin{figure}[!t]
\centering 
\subfigure[Cora-ML]{
\includegraphics[bb= 0 0 450 350,width=0.21\textwidth]{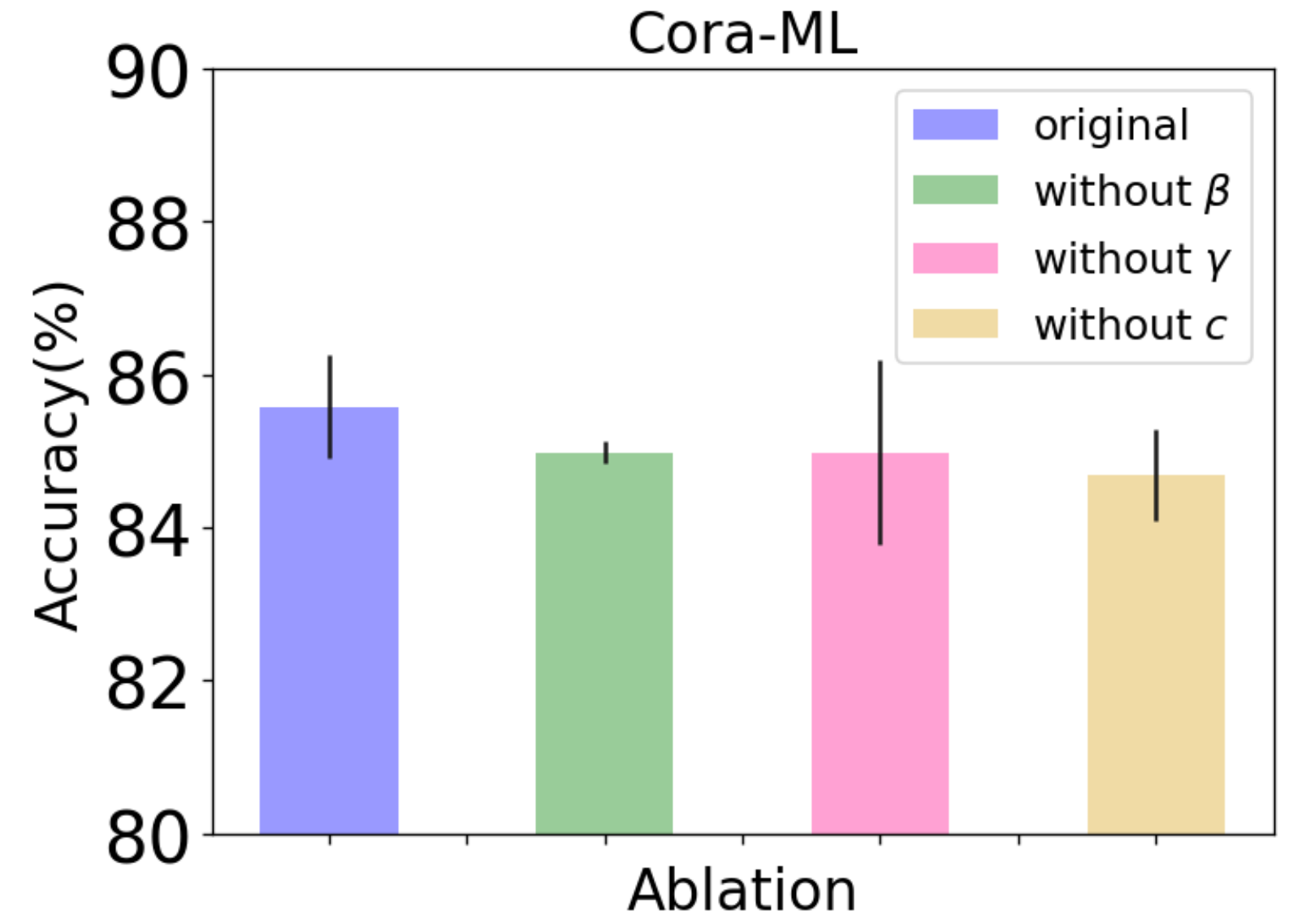}
}
\quad
\subfigure[Citeseer]{
\includegraphics[bb= 0 0 450 350,width=0.21\textwidth]{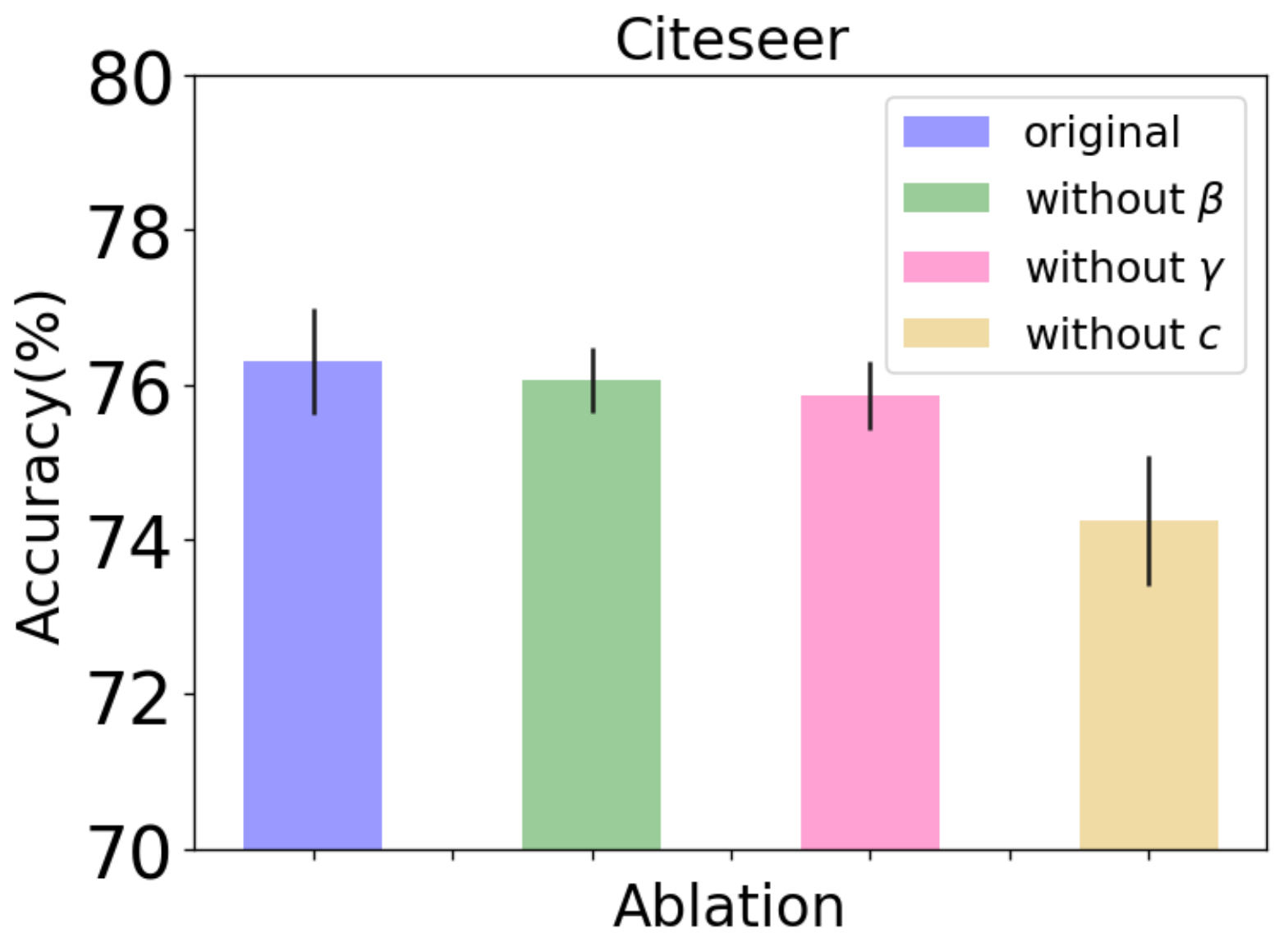}
}
\caption{Performance (Accuracy$\pm$Std) without each term}\label{fig:ablation_study}
\end{figure}

In Eq.~\eqref{eq:lower_opt_fast}, the term $ \beta\left(\|\mathbf{p}\|^{2}+\|\mathbf{q}\|^{2}\right)$ is the norm term that measures the propagation matrix difference $(T1)$, the term $\gamma \operatorname{tr}(\mathbf{X}^{\top} \mathbf{p q}^{\top} \mathbf{X})$ is the feature term $(T2)$. The term 
$\sum_{a_{i}, a_{j} \in \mathcal{C}_{k} \atop b_{j} \in \mathcal{V}^{(y)} \backslash \mathcal{C}_{k}} c g(\mathbf{Q}_{s}(a_{i}, b_{j})-\mathbf{Q}_{s}(a_{i}, a_{j}))$ reduces the discrepancy between the graph topology and the label distribution $(T3)$. Parameters $\beta$, $\gamma$ and $c$ control the influence of these terms on the lower level loss function. In the ablation study, we set one of them to zero at a time and observe how the performance changes on Cora-ML and Citeseer datasets. The mean accuracy and standard deviation are reported in Figure~\ref{fig:ablation_study}. 

When we remove $T1$, the accuracy decreases, which means if the learned PPR propagation matrix deviates too far from the original, the model performance will deteriorate. When $T2$ is removed, the accuracy decreases, which means the feature distribution plays a significant role in training. When we delete $T3$, it is equivalent to just learn the PPR propagation matrix with the feature distribution and controlling the deviation from the original $\mathbf{Q}$, without considering downstream tasks, which leads to the accuracy rate drop dramatically. Therefore, it is essential to find a balance among these terms. In this and Subsection \ref{subsec:parameter_analysis}, we only adopt one of the five data segmentation seeds mentioned in Subsection~\ref{subsec:effectiveness}, since experiments with different data segmentation seeds show similar trends. 

\subsection{Hyperparameter Analysis} \label{subsec:parameter_analysis}
\begin{figure}[!t]
\centering
\subfigure{
\includegraphics[bb= 0 0 500 400,width=3.8cm]{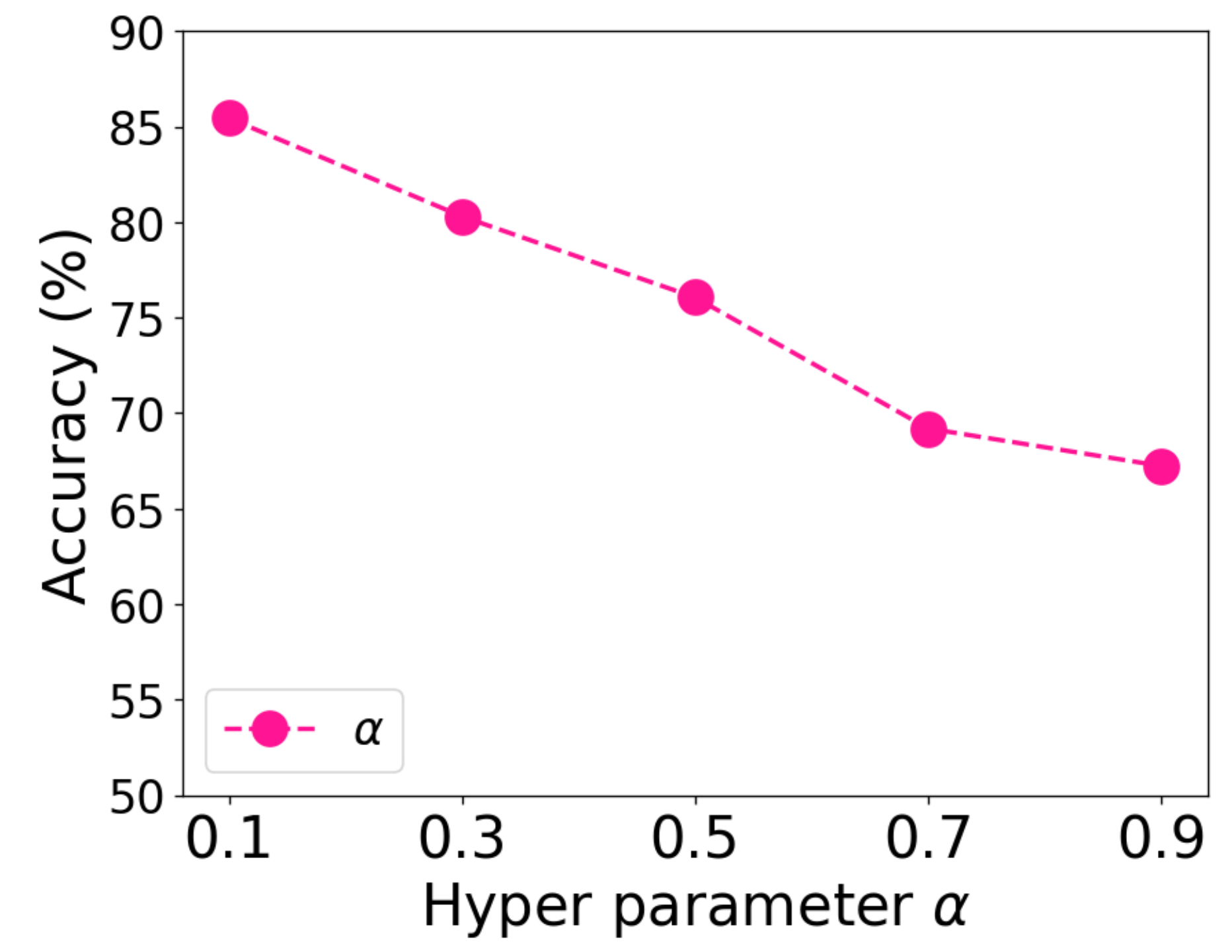}
}
\quad
\subfigure{
\includegraphics[bb= 0 0 500 400,width=3.8cm]{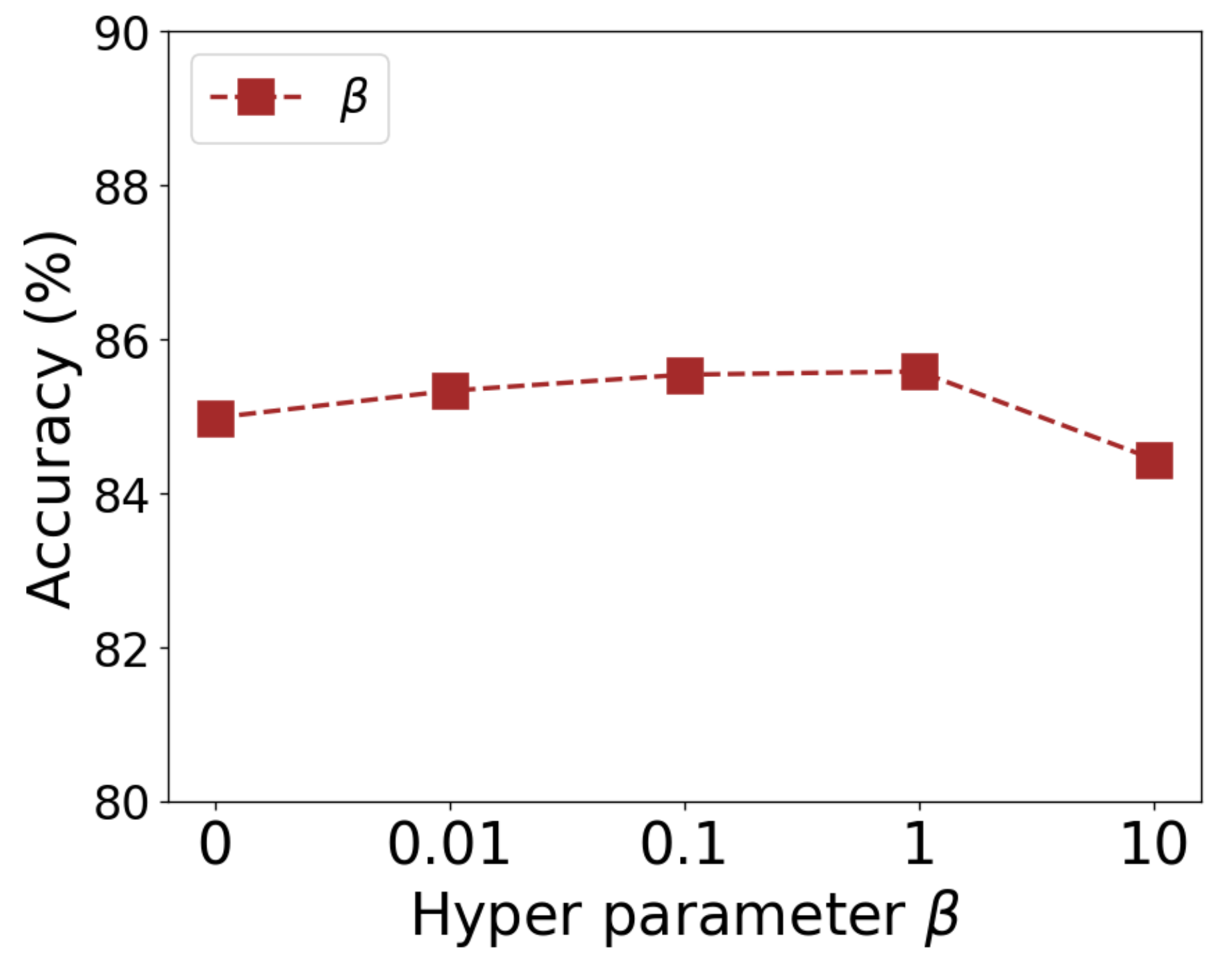}
}
\quad
\subfigure{
\includegraphics[bb= 0 0 500 400,width=3.8cm]{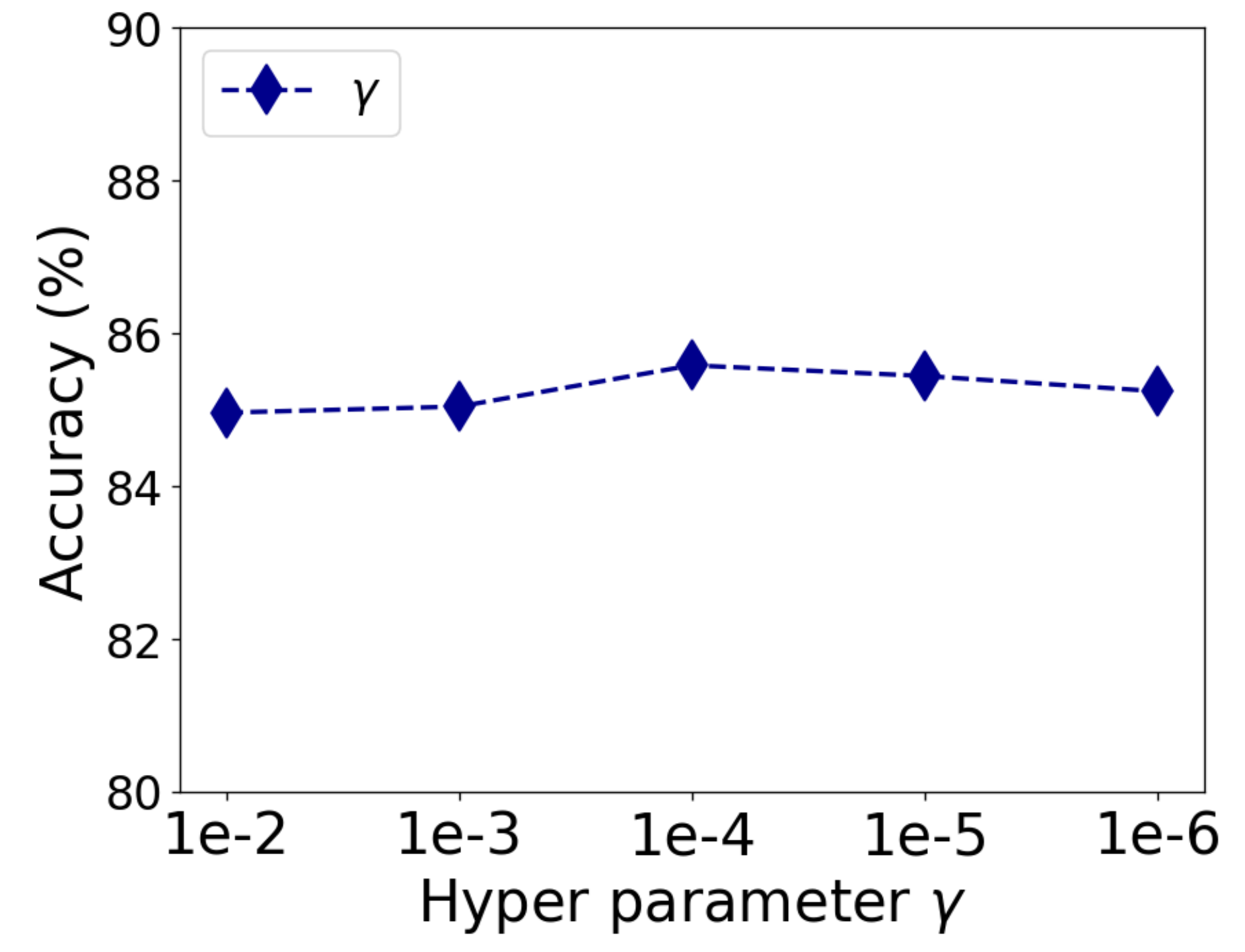}
}
\quad
\subfigure{
\includegraphics[bb= 0 0 500 400,width=3.8cm]{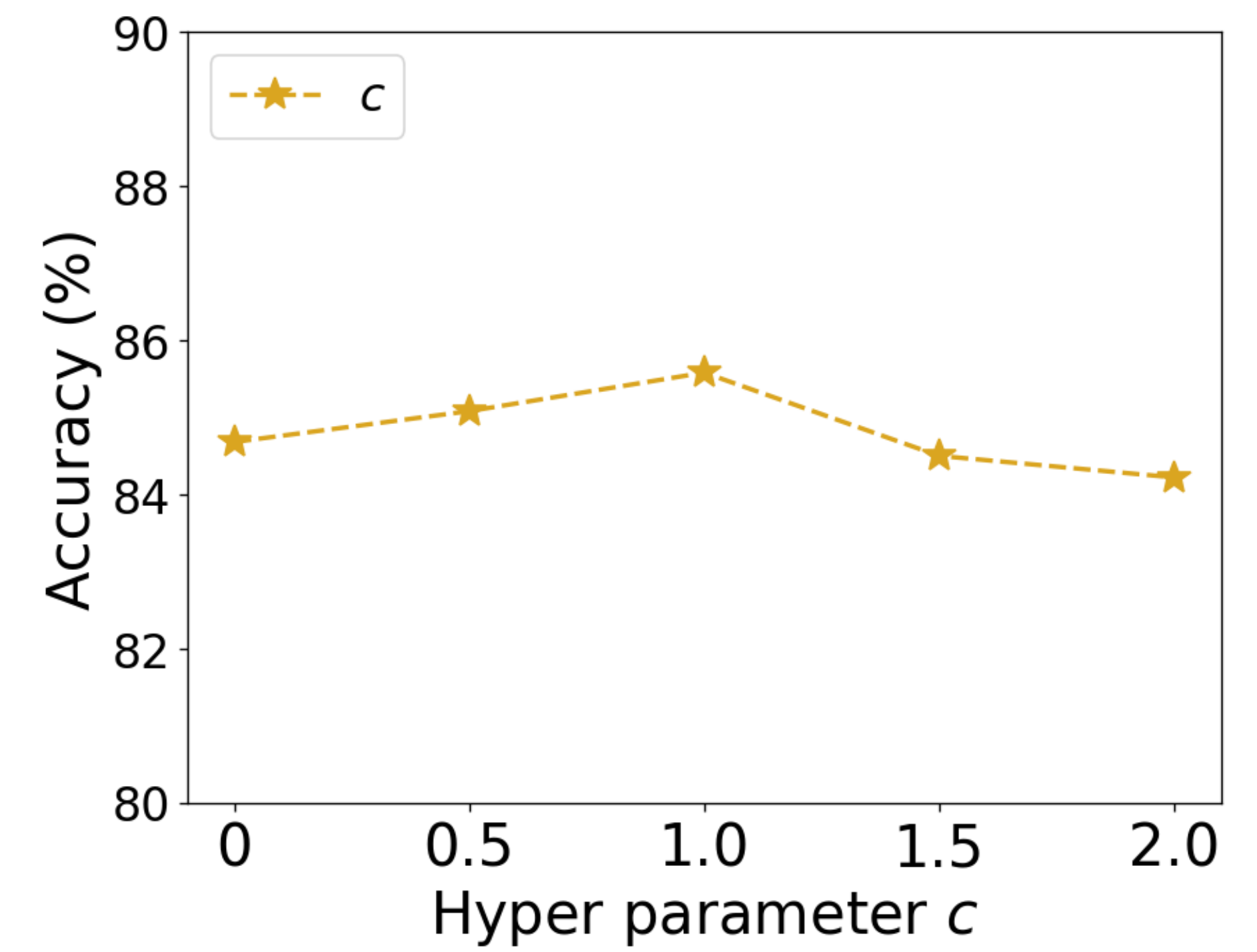}
}
\caption{Parameter analysis of A\optgnn\ on Cora-ML dataset}\label{fig:para_analysis}
\end{figure}

In this section, we discuss the sensitivity to hyperparameters $\alpha, \beta, \gamma$ and $c$ of our model by adjusting one parameter and keeping others constant.  Specifically, $\alpha$ ranges from $0.1$ to $0.9$ with interval $0.1$, $\beta$ ranges from $0.01$ to $10$ in a log scale of base $10$, $\gamma$ ranges from $1e-7$ to $1e-2$  in a log scale of base $10$ and $c$ ranges from $0.5$ to $2$ with interval $0.5$. When  $\alpha=0.1, \beta=1, \gamma=1e-4$ and $c=1$, our model performs best on Cora-ML. Due to space, we do not show figures of parameter analysis on other datasets, which are similar to Figure~\ref{fig:para_analysis}.
When A\optgnn\ achieves the best performance on Citeseer and Polblogs, parameters are $\alpha=0.1, \beta=1, \gamma=1e-5$, $c=1$ and $\alpha=0.1, \beta=1, \gamma=1e-5$, $c=2$. We can also see that in the tested range of hyperparameters, $\beta, \gamma, c$ show stable performance, while the performance drops as $\alpha$ grows larger. Supplementary explanation is given for that the trend of $\alpha$ is differnt from those of $\beta, \gamma$ and $c$. In Figure~\ref{fig:para_analysis}, although $\alpha$, $\beta$, $\gamma$ and $c$ are all parameters that control the performance of the model,they perform distinct roles. $\beta$, $\gamma$ and $c$ are trade-off parameters of Eq.~\eqref{eq:lower_opt_fast}, while $\alpha$ is a parameter of PPR that controls teleport (or restart) probability. Therefore, the trends of parameters in Figure~\ref{fig:para_analysis} are different. 

\subsection{Efficiency Comparison}\label{sec:time_comparison}
\begin{figure}[!t]
\centering
\includegraphics[bb= 0 0 400 300,width=0.35\textwidth]{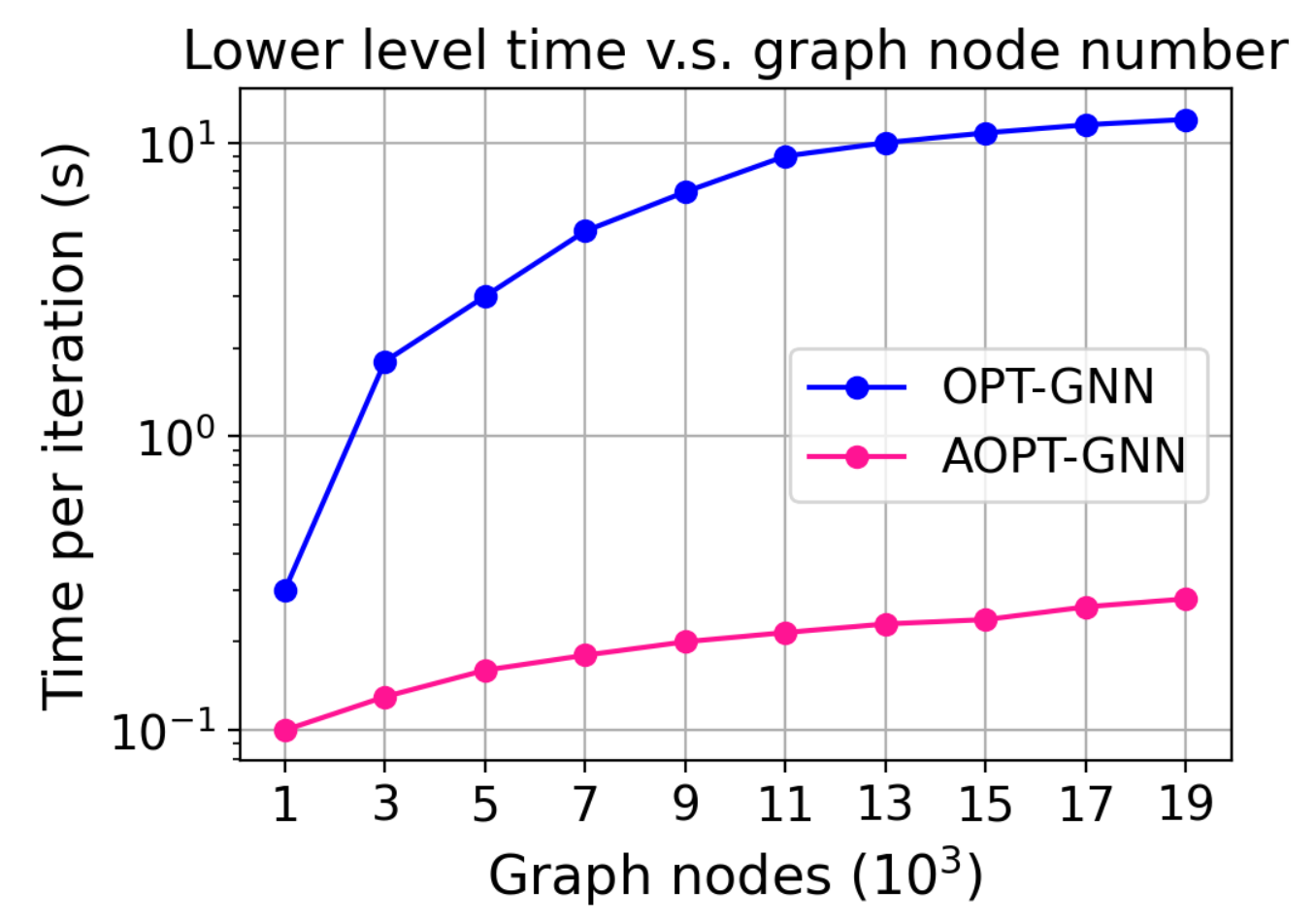}
\caption{Lower level time v.s. graph node number}
\label{fig:time_comparison}
\end{figure}

We conduct experiments to compare the running time of AOPT-GNN and OPT-GNN. The difference in efficiency between the two methods lies in the lower-level optimization, so we measure these two methods' lower-level training time on the graph with different number of nodes. The results is presented in Figure~\ref{fig:time_comparison}. As we can observe, the AOPT-GNN significantly improves the efficiency compared to OPT-GNN. 

\section{Related work}\label{sec:related-work}
\noindent \textbf{PPR Propagation Matrix Based Models.}
Some of existing PPR propagation matrix-based methods are based on the Monte Carlo method \cite{wang2017fora,wei2018topppr} which simulate random walks from root nodes. 
GDC \cite{klicpera2019diffusion} approximates the distribution with Poisson distribution. It uses symmetric normalization adjacency matrix and changes the one-hot vector to the teleporting probability distribution. 
SGC \cite{wu2019simplifying} considers improving the structure of GCN from the perspective of PPR propagation matrix by multiplication of symmetric normalized adjacency matrix layers instead of convolution to construct a linear structure, which significantly speeds up the propagation without compromising accuracy. 
AGP \cite{wang2021approximate} unifies above methods by adding variables to weights and propagation matrices. The randomized algorithm makes the computational complexity independent of the graph size. 
GPR-GNN \cite{chien2020adaptive} combines generalized PageRank (GPR) scheme with GNNs and adaptively learns GPR weights, realizing joint node feature optimization and topological information extraction. Also, it mitigates feature over-smoothing.
Compared to the vanilla GCN which uses a full-batch training, these PPR propagation matrix-based methods decouple prediction and propagation, and thus allow mini-batch training to improve the scalability.

\noindent \textbf{Graph Data Augmentation.}
Graph data augmentation has been proposed to solve a variety of graph tasks \cite{ding2022data}. One of the latest work \cite{xu2022graph} applies bi-level optimization to graph sanitation problems by training a backbone classifier in the lower level, and modifying the initial graph to improve the performance with a downstream classifier. \cite{roy2021node} combines the local aggregation and non-local aggregation to generate node representations with essential features that are `washed out' local aggregation. \cite{franceschi2019learning,jin2020graph}  simultaneously learn the graph structure and parameters of a GNN, which improves the robustness under various adversarial attacks, real-world noise and incompleteness of graphs. 
TO-GNN \cite{yang2019topology} uses gradient descent-based graph updates and smoothness-related regularizations, focusing on optimizing graph-related distributions, such as graph generation and edge dropping, to facilitate graph rewiring through sampling. \cite{song2021topological} proposes another graph topological regularization by adding unsupervised learning representation as additional features to the graph. 
However, most existing methods aim to learn an optimal adjacency matrix that requires multiple propagation steps, and is limited by the sparsity property. Instead, we directly learn the optimal propagation matrix which has the advantages of efficiency and flexibility. 

\section{Conclusion}\label{sec:conclusion}
In this paper, we propose a bi-level programming-based approach which directly learns an optimal propagation matrix as an implicit modification of the graph topology, and also improves the robustness when facing attacks. We further propose a low-rank model to speed up the computation of model training. Experimental results show that the two proposed models outperform all the representative baselines in node classification. The adversarial attack evaluation demonstrates the strong robustness of the proposed models under attacks.

One limitation of this work is that it is under the transductive graph learning setting, if additional nodes are added to the graph, the propagation matrix may no longer be optimal. In this case, if the number of new nodes is small enough, the updated propagation might still have better performance than using the original adjacency and propagation matrix. How to extend propagation matrix optimization to the inductive setting is one future direction.


\bibliography{LaTeX/aaai24}



\end{document}